\documentclass[letterpaper, 10 pt, conference]{ieeeconf}  

\IEEEoverridecommandlockouts                              
\usepackage{amsmath,amssymb,mathrsfs}
\usepackage[ruled, linesnumbered, vlined]{algorithm2e}
\usepackage{comment}
\usepackage{graphicx}
\usepackage{graphics}
\usepackage[tight]{subfigure}
\usepackage{float}
\usepackage[table]{xcolor}
\usepackage{multirow}
\usepackage{array}
\usepackage{enumerate}
\usepackage{sidecap}
\usepackage{hyperref}
\usepackage{url}
\usepackage{wrapfig}
\usepackage{bm}
\usepackage{algpseudocode}
\graphicspath{{./figs/}}
\usepackage{flushend}

\usepackage[space]{cite}

\usepackage{color}

\definecolor{lightblue}{rgb}{0,0.2,1}
\definecolor{black}{rgb}{0,0,0}
\hypersetup{
    colorlinks=true,%
    citebordercolor=white,%
    filebordercolor=white,%
    linkbordercolor=white,%
    urlbordercolor=white,%
    citecolor=black,%
    linkcolor=black,%
    urlcolor=black%
}

\newcounter{tecounter}
\DeclareMathOperator*{\argmin}{arg\,min}
\DeclareMathOperator*{\argmax}{arg\,max}

\title{\LARGE \bf
Informative Planning and Online Learning\\  with Sparse Gaussian Processes 
}

\author{Kai-Chieh Ma, Lantao Liu, Gaurav S. Sukhatme
\thanks{
The authors are with the Department of Computer Science at the
University of Southern California, Los Angeles, CA 90089, USA.
        {\tt\small \{kaichiem, lantao.liu, gaurav\}@usc.edu}}%
}

\begin{document}

\maketitle
\thispagestyle{empty}
\pagestyle{empty}

\begin{abstract}
A big challenge in environmental monitoring is the spatiotemporal variation of the phenomena to be observed. 
To enable persistent sensing and estimation in such a setting, it is beneficial to have a time-varying
underlying environmental model.
Here we present a planning and learning method that enables an autonomous marine vehicle to perform persistent ocean monitoring tasks by learning and refining an environmental model. 
To alleviate the computational bottleneck caused by large-scale data accumulated, we propose a framework that iterates between a planning component aimed at collecting the most information-rich data, and a sparse Gaussian Process learning component where the environmental model and hyperparameters are learned online by taking advantage of only a subset of data that provides the greatest contribution. Our simulations with ground-truth ocean data shows that the proposed method is both accurate and efficient.
\end{abstract}


\section{Introduction and Related Work}

Scientists are able to gain a greater understanding of the environmental processes (e.g., physical, chemical or biological parameters) through environmental sensing and monitoring~\cite{Dunbabin12}.
However, many environmental monitoring scenarios involve large environmental space and require considerable amount of work for collecting the data. 
Increasingly, a variety of autonomous robotic systems including marine vehicles~\cite{Fiorelli03adaptivesampling}, aerial vehicles~\cite{classification12}, and ground vehicles~\cite{Trincavelli08}, are designed and deployed for environmental monitoring in order to replace the conventional method that deploys static sensors to areas of interest~\cite{Oliveira11}.
Particularly, the autonomous underwater vehicles (AUVs) such as marine gliders are becoming popular due to their long-range (hundreds of kilometers) and long-term (weeks even months) monitoring capabilities~\cite{Miles2015,PaleyLeoZhang2006,LPDDLZ_JFR10}.

\begin{figure}[t]
 \centering
  \subfigure[Aug 15, 2016]
        {\label{fig:815}\includegraphics[height=1.1in]{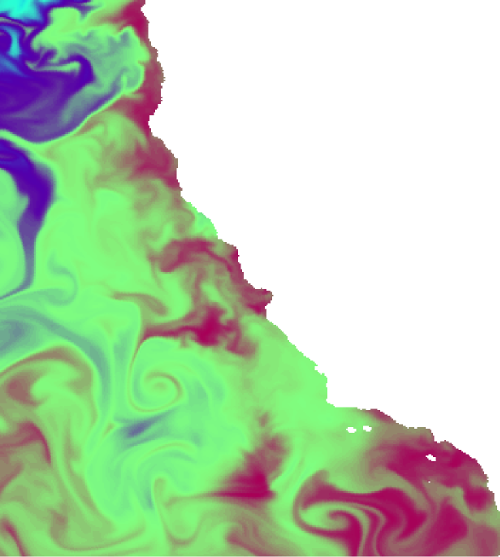}}
 \quad
 \subfigure[Aug 22, 2016]
        {\label{fig:822}\includegraphics[height=1.1in]{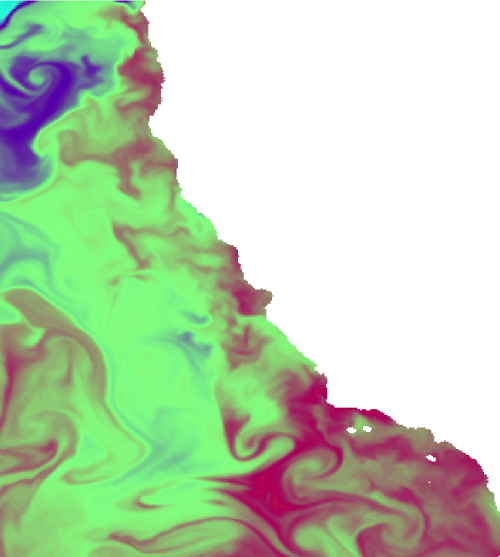}}
 \quad 
 \subfigure[Aug 29, 2016]
        {\label{fig:829}\includegraphics[height=1.1in]{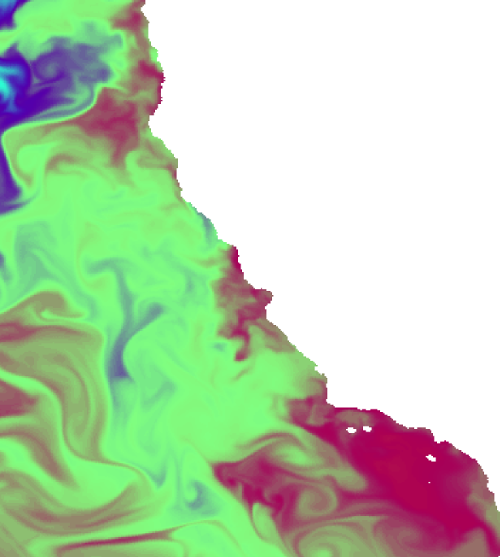}}
 \caption{Ocean salinity data in the Southern California Bight region generated by the Regional Ocean Modeling Systems (ROMS)~\cite{shchepetkin_regional_2005}. Color indicates levels of salinity content. 
 }
\label{fig:salinity} 
\end{figure}

We are interested in the problem of collecting data about a scalar field of important environmental attributes such as temperature, salinity, or chlorophyll content of the ocean, and learn a model to best describe the environment (i.e., levels or contents of the chosen attribute at every spot in the entire field). 
However, the unknown environmental phenomena that we are interested in can be non-stationary~\cite{Ouyang2014MAS}. Fig.~\ref{fig:salinity} shows the variations of salinity data in the Southern California Bight region generated by the Regional Ocean Modeling Systems (ROMS)~\cite{shchepetkin_regional_2005}.
In order to provide a good estimate of the state of the environment and maintain the prediction model at any time, the environmental sensing (information gathering) needs to be carried out persistently to catch up to possible variations~\cite{Meliou07}.

We aim at estimating the current state of the environment and providing a nowcast (not forecast or hindcast) of the environment, via navigating the robots to collect the information. 
To model spatial phenomena, a common approach is to use a rich class of Gaussian Processes~\cite{Rasmussen2005,Singh2007,Ouyang2014MAS} in spatial statistics.
In this work, we also employ this broadly-adopted approach to build and learn an underlying model of interest.

Still, there are challenges:
\begin{itemize}
\item 
The first challenge lies in the model learning with the most useful sensing inputs,
i.e., we wish to seek for the samples that best describe the environment.
Navigating the robot to obtain such samples is called {\em informative planning}~\cite{binney13}. 
In this work, we utilize the mutual information between visited locations and the remainder of the space to characterize the amount of information (information gain) collected. 

\item
The second challenge is the relaxation of the prohibitive computational cost for the model prediction. 
The most accurate way to estimate a latent model is to use all historical sensing data. 
However, since the environmental monitoring task can be long-range and long-term, the data size continuously grows until it ``explodes". 
Consequently, an efficient estimator will need to dynamically select only the most information-rich data while abandoning the samples that are less informatively novel.
    
\end{itemize}

Planning and environment monitoring are two big and well studied topics. Here we briefly review the works that are related to the informative planning and the model prediction with sparse GPs.
Representative informative planning approaches include, for example, algorithms based on a recursive-greedy style~\cite{Singh2007,Meliou07} where the informativeness is generalized as submodular functions and a sequential-allocation mechanism is designed in order to obtain subsequent waypoints. 
This recursive-greedy framework has been extended later by incorporating obstacle avoidance~\cite{Binney-2010-642} and diminishing returns~\cite{binney13}. 
In addition, a differential entropy based framework~\cite{Low2009thesis,Cao2013} was proposed where a batch of waypoints can be obtained through dynamic programming.
We recently proposed a similar informative planning method based on the dynamic programming structure in order to compute the informative waypoints~\cite{ma2016information}. This method is further extended here as an adaptive path planning component by incorporating the online learning and re-planning mechanisms.
There are also many methods optimizing over complex deterministic and static information (e.g., see~\cite{SolteroSR12,YuSchRus14ICRA}). 

A critical problem for the persistent (long-term even life-long) tasks that one must consider is the large-scale accumulated data. Although affluent data might predict the most accurate model, in practice a huge amount of data are very likely to exceed the capacity of onboard computational hardware. 
Methods for reducing the computing burdens of GPs have been proposed. For example, GP regressions can be done in a real-time fashion where the problem can be estimated locally with local data~\cite{Nguyen-tuong08localgaussian}.
Another representative framework is a sparse representations of the GP model~\cite{csato2002sparse} which is based on a combination of a Bayesian online algorithm together with a sequential construction of the most relevant subset of the data. 
This method allows the model to be refined in a recursive way as the data streams in. 
The framework has been further extended to many application domains such as visual tracking~\cite{Ranganathan11}. 

We propose an informative planning and online learning approach for the long-term environmental monitoring. The objective is to construct an estimated model by navigating the robot to the most informative regions to collect data with the greatest information. Our method integrates the sparse variant of GPs so that both the model and the hyperparameters can be improved online with dynamic but a fixed size of data. Then the ameliorated environment model is in turn used to improve the planning component at appropriate re-planning moments. 
We conducted simulation on ocean temperature data and the results show that the predicted model can very well match the patterns of the ground truth model.


\section{Preliminaries}

In this section, we briefly present the preliminary background for the GP-based environmental modeling.

\subsection{Gaussian Process Regression on Spatial Data}

A GP is defined as a collection of random variables where any finite number of which have a joint Gaussian distribution. 
GP's prediction behavior is determined by the prior covariance function (also known as {\em kernel}) and the training points.
The prior covariance function describes the relation between two independent data points and it typically comes with some free hyperparameters to control the relation.

Formally, let $X$ be the set of $n$ training points associated with target values, $\bm{y}$,  and let $X_*$ be the testing points. The predictive equations of the GP regression can be summarized as:
\begin{equation} \label{eq:gp}
    \begin{aligned}
        \bm{f}_* | X, \bm{y}, X_* &\sim \mathcal{N}(\bar{\bm{f}_*}, cov(\bm{f}_*)) \\
        \bar{\bm{f}_*} &\triangleq \mathbb{E}[\bm{f}_* | X, \bm{y}, X_*] = K(X_*, X)K(X, X)^{-1} \bm{y} \\
		cov(\bm{f}_*) &= K(X_*, X_*) - K(X_*, X)K(X, X)^{-1}K(X, X_*)
	\end{aligned}
\end{equation}
where $K(\cdot, \cdot)$ denotes a covariance matrix. For example, $K(X, X_*)$ is evaluated by a pre-selected kernel function for all pairwise data points in $X$ and $X_*$. 
A widely adopted choice of kernel function for spatial data is the {\em squared exponential automatic relevance determination} function: 
\begin{equation}
	\begin{split}
	k(\bm{x}, \bm{x}') = \sigma_f^2 \exp(-\frac{1}{2}(\bm{x}-\bm{x}')^TM(\bm{x}-\bm{x}')) + \sigma_n^2\delta_{xx'} 
	\end{split}
\end{equation}
where
$M = diag(\bm{l})^{-2}$. The parameters $\bm{l}$ are the {\em length-scales} in each dimension of $\bm{x}$ and determine the level of correlation (each $l_i$ models the degree of smoothness in the spatial variation of the measurements in the $i$th dimension of the feature vector $\bm{x}$). $\sigma_f^2$ and $\sigma_n^2$ denote the variances of the signal and noise, respectively. $\delta_{xx'}$ is the Kronecker delta function which is 1 if $\bm{x}=\bm{x}'$ and zero otherwise.

\subsection{Estimation of Hyperparameters Using Training Data}\label{sec:HyperTraining}
Let $\bm{\theta} \triangleq \{\sigma_n^2, \sigma_f^2, \bm{l}\}$ be the set of hyperparameters in the kernel function. We are interested in estimating these hyperparameters so that the kernel function can describe the underlying phenomena as accurate as possible.
A common approach to learning the set of hyperparameters is via maximum likelihood estimation combined with $k$-fold cross-validation (CV)~\cite{Rasmussen2005}. An extreme case of the $k$-fold cross-validation is when $k = n$, the number of training points, also known as leave-one-out cross-validation (LOO-CV). Mathematically, the log-likelihood when leaving out training case $i$ is
\begin{equation}
	\log p(y_i | X, \bm{y}_{-i}, \bm{\theta}) = -\frac{1}{2}\log\sigma_i^2 - \frac{(y_i - \mu_i)^2}{2\sigma_i^2} - \frac{1}{2}\log(2\pi)
\end{equation}
where $\bm{y}_{-i}$ denotes all targets in the training set except the one with index $i$, and $\mu_i$ and $\sigma_i^2$ are calculated according to Eq.~\eqref{eq:gp}. The log-likelihood of LOO is therefore
\begin{equation}
	L_{LOO}(X, \bm{y}, \bm{\theta}) = \sum_{i = 1}^{n}\log p(y_i | X, \bm{y}_{-i}, \bm{\theta}).
\end{equation}
Notice that in each of the $|\bm{y}|$ LOO-CV iterations, a matrix inverse, $K^{-1}$, is needed, which is costly if computed repeatedly. This can actually be computed efficiently from the inverse of the complete covariance matrix using {\em inversion by partitioning}~\cite{press1996numerical}. The resulting predictive mean and variance can then be formulated as
\begin{equation} 
    \begin{aligned}
        \mu_i &= y_i - [K^{-1}\bm{y}]_i / [K^{-1}]_{ii} \\
        \sigma_i^2 &= 1 / [K^{-1}]_{ii}
	\end{aligned}
\end{equation}

To obtain the optimal values of hyperparameters $\bm{\theta}$, we can compute the partial derivatives of $L_{LOO}$ and use the conjugate gradient optimization techniques.
The partial derivatives of $L_{LOO}$ is
\begin{equation} 
    \begin{aligned}
        \frac{\partial L_{LOO}}{\partial \theta_j} = \sum_{i = 1}^{n} &\frac{1}{[K^{-1}]_{ii}} \Big(\alpha_i[Z_j\bm{\alpha}]_i \\ &-\frac{1}{2}(1 + \frac{\alpha_i^2}{[K^{-1}]_{ii}}) [Z_j K^{-1}]_{ii}\Big),
	\end{aligned}
\end{equation}
where $\bm{\alpha} = K^{-1}\bm{y}$ and $Z_j = K^{-1}\frac{\partial K}{\partial \theta_j}$.
With the standard gradient descent method, we update each $\theta_j$ iteratively:
\begin{equation} 
    \theta_j^{(t+1)} = \theta_j^{(t)} + \eta \frac{\partial L_{LOO}}{\partial \theta_j^{(t)}},
\end{equation}
where $\eta$ is the learning rate.


\section{Technical Approach}

As aforementioned, one limitation of GPs for the long-term mission is the memory requirement for large (possibly infinite) training sets. 
In our system, we borrow the idea of Sparse Online Gaussian Process (SOGP)~\cite{csato2002sparse} to overcome this limitation. The method is based on a combination of a Bayesian online algorithm together with a sequential construction of a relevant subsampling of the data which best describes a latent model.

\subsection{Online Learning with Gaussian Processes} \label{sec:OnlineGP}
Given a prior GP $\hat{p}_t(\bm{f})$ at time $t$, when a new data point $(\bm{x}_{t+1}, y_{t+1})$ at time $t+1$ comes in, it's incorporated by performing a Bayesian update to yield a posterior.
\begin{equation}
	p_{post}(\bm{f}) = \frac{p(y_{t+1}|\bm{f}) \hat{p}_t(\bm{f})}{\mathbb{E}_{\hat{p}_t(\bm{f})}[p(y_{t+1}|\bm{f}_{D})]},
\end{equation}
where $\bm{f} = [f(\bm{x}_1), \dots, f(\bm{x}_M)]^{T}$ denotes a set of function values, and $\bm{f}_{D} \subseteq \bm{f}$ where $\bm{f}_{D}$ is the set of $f(\bm{x}_i) = f_i$ with $\bm{x}_i$ in the training set. In general, $p_{post}(\bm{f})$ is no longer Gaussian unless the likelihood itself is also Gaussian. Therefore, $p_{post}(\bm{f})$ is projected onto the closest GP, $\hat{p}_{t+1}$ where $\hat{p}_{t+1} = \argmin_{\hat{p}}$ KL$(p_{post}(\bm{f}) || \hat{p})$.
(KL is the Kullback-Leibler divergence that is used to measure the difference between two probability distributions.) 
It is shown in ~\cite{Opper:1999:BAO:304710.304756} that the projection results in a good matching of the first two moments (mean and covariance) of $p_{post}$ and the new Gaussian posterior $\hat{p}_{t+1}$. By following the lemma of ~\cite{Csató_sparseon-line}, we arrive at the parametrization for the approximate posterior GP at time $t$ as a function of the kernel and likelihoods (“natural parametrization”):
\begin{equation} \label{eq:SOGP_Posterior}
    \begin{aligned}
        \bar{f_*} &= \sum_{i=1}^{t} k(\bm{x}_*, \bm{x}_i) \alpha_t(i) = \bm{\alpha_t}^T\bm{k}_{\bm{x}_*, t} \\
        var(f_*) &= k(\bm{x}_*, \bm{x}_*) + \sum_{i, j = 1}^{t} k(\bm{x}_*, \bm{x}_i)[C_t]_{ij}k(\bm{x}_j, \bm{x}_*) \\
                &= k(\bm{x}_*, \bm{x}_*) + \bm{k}_{\bm{x}_*, t}^T C_t \bm{k}_{\bm{x}_*, t}
	\end{aligned}
\end{equation}
where $\bm{k}_{\bm{x}_*, t} = [k(\bm{x}_1, \bm{x}_*), \dots, k(\bm{x}_t, \bm{x}_*)]^T$, and $\bm{\alpha_t}$ and $C_t$ are updated using
\begin{equation}
    \begin{aligned}
        \bm{\alpha_t} &= T_t(\bm{\alpha_{t-1}}) + q_t \bm{s}_t \\
        C_t &= U_t(C_{t-1}) + r_t \bm{s}_t \bm{s}_t^T \\
        \bm{s}_t &= T_t(C_{t-1} \bm{k}_{\bm{x}_*, t}) + \bm{e}_{t} \\
        q_t &= \frac{\partial}{\partial\mathbb{E}_{\hat{p}_{t-1}(\bm{f})}[\bm{f}_t]} \log\mathbb{E}_{\hat{p}_{t-1}(\bm{f})}[p(y_t | \bm{f}_t)] \\
        r_t &= \frac{\partial^2}{\partial\mathbb{E}_{\hat{p}_{t-1}(\bm{f})}[\bm{f}_t]^2} \log\mathbb{E}_{\hat{p}_{t-1}(\bm{f})}[p(y_t | \bm{f}_t)] \\
	\end{aligned}
\end{equation}
where $\bm{e}_{t}$ is the $t$-th unit vector. The operator $T_t$ ($U_t$) is defined to extend a $t-1$-dimensional vector (matrix) to a $t$-dimensional one by appending zero at the end of the vector (zeros at the last row and column of the matrix).
For the regression with Gaussian noise (variance $\sigma_0^2$), The expected likelihood is a normal distribution with mean $\bar{f_*}$ and variance $var(f_*) + \sigma_0^2$. Hence, the logarithm of the expected likelihood is:
\begin{equation}
  \begin{split}
    \log\mathbb{E}_{\hat{p}_{t-1}(\bm{f})}[p(y_t | \bm{f}_t)] &= -\frac{1}{2}\log[2\pi(var(f_*) + \sigma_0^2)] \\
    &- \frac{(y_t - \bar{f_*})^2}{2(var(f_*) + \sigma_0^2)},
  \end{split}
\end{equation}
and the first and second derivatives with respect to the mean $\bar{f_*}$ give the scalars $q_t$ and $r_t$ are
\begin{equation}
    \begin{aligned}
        q_t &= \frac{y_t - \bar{f_*}}{var(f_*) + \sigma_0^2}, \\
        r_t &= -\frac{1}{var(f_*) + \sigma_0^2}. \\
	\end{aligned}
\end{equation}


\subsection{Sparseness in Gaussian Processes} \label{sec:SparseGP}

To prevent the unbounded growth of memory requirement due to the increase of data, it is necessary to limit the number of the training points which are stored in a {\em basis vector set (BV-set)},  while preserving the predictive accuracy of the model. This is done in two different stages. 

First, when the new training point $(\bm{x}_{t+1}, y_{t+1})$ at time $t+1$ arrives, we calculate the squared norm of the ``residual vector" from the projection in the space spanned by the current BV-set. Let the quantity be $\gamma_{t+1}$, specifically, 
\begin{equation}
    \gamma_{t+1} = k(\bm{x}_{t+1}, \bm{x}_{t+1}) - \bm{k}_{\bm{x}_{t+1}, t}^T Q_{t} \bm{k}_{\bm{x}_{t+1}, t},
\end{equation}
where $Q_{t} = K(X_{t}, X_{t})^{-1}$ is the inversion of the full kernel matrix.
The costly matrix inversion can be alleviated via the following equations:
\begin{equation}
    \begin{aligned}
        Q_t &= U_t(Q_{t-1}) + \gamma_t^{-1}(T_t(\hat{\bm{e}}_{t}) - \bm{e}_{t})(T_t(\hat{\bm{e}}_{t}) - \bm{e}_{t})^T \\
        \hat{\bm{e}}_{t} &= Q_{t-1}^{-1} \bm{k}_{\bm{x}_t, t-1}
    \end{aligned}
\end{equation}
Essentially, $\gamma_{t+1}$ can also be thought as a form of ``novelty" for the new training point $(\bm{x}_{t+1}, y_{t+1})$. Therefore it's included in BV-set only if it exceeds some predefined threshold $\omega$. 
Otherwise, only an update of $\hat{\bm{s}}_{t+1}$ is necessary.

\begin{equation}
    \hat{\bm{s}}_{t+1} = C_t \bm{k}_{\bm{x}_{t+1}, t} + \hat{\bm{e}}_{t+1}
\end{equation}

Second, when the size of BV-set exceeds the memory limit (or any pre-defined limit), $m$, a score measure is used to pick out the lowest one and remove it from the existing BV-set. Formally, let $\epsilon_i$ be the scoring function for the $i$th element in the BV-set. It's a measure of change on the expected posterior mean of a sample due to sparse approximation~\cite{Csató_sparseon-line}. 
\begin{equation}
    \epsilon_i = \frac{|[\bm{\alpha}_{t+1}]_i|}{[Q_{t+1}]_{ii}}
\end{equation}
Assume the $j$th element in BV-set is the one with the lowest $\epsilon$, the removal of any element requires a re-update of parameters $\bm{\alpha}_{t+1}$, $C_{t+1}$ and $Q_{t+1}$.
\begin{equation}
    \begin{aligned}
        \hat{\bm{\alpha}}_{t+1} &= \bm{\alpha}^{(t)} - \alpha^j \frac{Q^j}{q^j} \\
        \hat{C}_{t+1} &= C^{(t)} + c^j\frac{Q^jQ^{jT}}{q^{j2}} - \frac{1}{q^j}[Q^jC^{jT} + C^{j}Q^{jT}] \\
        \hat{Q}_{t+1} &= Q^{(t)} - \frac{Q^{j} Q^{jT}}{q^j},
	\end{aligned}
\end{equation}
where $C^{(t)}$ is the resized matrix by removing the $j$th column and the $j$th row from
$C_{t+1}$, $C^j$ is the $j$th column of $C_{t+1}$ excluding the $j$th element and $c^j = [C_{t+1}]_{jj}$. 
Similar operations apply for $Q^{(t)}$, $Q^{j}$, $q^{j}$, $\bm{\alpha}^{(t)}$,  and $\alpha^j$.


\subsection{Environment Representation \& Informative Sampling Locations} \label{sec:infoPlanner}

To facilitate the computation of future informative sampling locations, we discretize the environment into a grid map where each grid represents a possible sampling spot. 
The mean and variance of the measurement at each grid can be predicted via the SOGP model. 
We use the {\em mutual information} between the visited locations and the remainder of the space to characterize the
amount of information (information gain) collected.
Formally, the mutual information between two sets of sampling spots, $A$, $B$ can be evaluated as:
\begin{equation}
    I(Z_A;Z_B) = I(Z_B; Z_A)= H(Z_A) - H(Z_A|Z_B).
\end{equation}
The entropy $H(Z_A)$ and conditional entropy $H(Z_A|Z_B)$ can be calculated by 
\begin{equation}
    \begin{aligned}
        H(Z_A) = \frac{1}{2}\log\Big((2 \pi e)^k |\Sigma_{AA}|\Big) \\
	    H(Z_A|Z_B) = \frac{1}{2}\log\Big((2 \pi e)^k|\Sigma_{A|B}|\Big) \\
	\end{aligned}
\end{equation}
where $k$ is the size of $A$. The covariance matrix $\Sigma_{AA}$ and $\Sigma_{A|B}$ can essentially be calculated from the posterior GP described in Eq.~\eqref{eq:SOGP_Posterior}.

To compute the future sampling spots, let $X$ denote the entire sampling space (all grids), and $Z_{X}$ be measurements for data points in $X$.
The objective is to find a subset of sampling points, $P\subset X$ with a size $|P| = n$, which gives us the most information for predicting our model.
This is equivalent to the problem of finding new sampling points in the un-sampled space that maximize the mutual information between sampled locations and un-sampled part of the map.
The optimal subset of sampling points, $P^*$, with maximal mutual information is
\begin{equation}\label{eq:objective}
	P^* = \argmax_{P\in\mathcal{X}}I(Z_P;Z_{X \setminus P})
\end{equation}
where $\mathcal{X}$ represents all possible combinatorial sets, each of which is of size $n$. $P^*$ can be computed efficiently using a dynamic programming (DP) scheme~\cite{ma2016information}. Here is the basic idea: 
Let $\bm{x}_i \in X$ denote an arbitrary sampling point at DP stage $i$ and $\bm{x}_{a:b}$ represent a sequence of sampling points from stage $a$ to stage $b$. 
The mutual information between the desired sampling points (which eventually form $P$) and the remaining map can then be written as $I(Z_{\bm{x}_{1:n}};Z_{X \setminus \{\bm{x}_{1:n}\}})$, which can be approximated as follows:
\begin{equation} \label{eq:mutual}
	\begin{split}
		I(Z_{\bm{x}_{1:n}}; &Z_{X \setminus \{\bm{x}_{1:n}\}}) \approx I(Z_{\bm{x}_1}; Z_{X \setminus \{\bm{x}_1\}}) \\
		 &+ \sum_{i=2}^{n}I(Z_{\bm{x}_i}; Z_{X \setminus \{\bm{x}_1, \dots, \bm{x}_i\}}|Z_{\bm{x}_{1:i-1}}),
	\end{split}
\end{equation}
Eq.~\eqref{eq:mutual} can be expressed in a recursive form, i.e. for stages $i = 2, \dots, n$, the value $V_i(\bm{x}_i)$ of $\bm{x}_i$ is:
\begin{equation*}
    \begin{split}
        V_i(\bm{x}_i) = &\max_{\bm{x}_i \in X \setminus \{ \bm{x}_1, \dots, \bm{x}_{i-1}\}}I(Z_{\bm{x}_i}; Z_{X \setminus \{ \bm{x}_1, \dots, \bm{x}_i\}}|Z_{\bm{x}_{1:i-1}}) \\
           & + V_{i-1}(\bm{x}_{i-1}),
    \end{split}
\end{equation*}
with a recursion base case 
$
	V_1(\bm{x}_1) = I(Z_{\bm{x}_{1}}; Z_{X \setminus \{\bm{x}_1\}}).
$
Then with the optimal solution in the last stage, $\bm{x}_{n}^* = \argmax_{\bm{x}_{n} \in X}V_{n}(\bm{x}_{n})$, we can backtrace all optimal sampling points until the first stage $\bm{x}_1^*$, and obtain $P^*=\{\bm{x}^*_{1}, \bm{x}^*_{2}, \dots , \bm{x}^*_{n}\}$.

Note that, the informativeness maximization procedure only outputs batches of sampling points, but does not convey any information of ``a path" which is a sequence of ordered waypoints. 
Therefore, these sampling points are post-processed with a customized Travelling Salesman Problem (TSP)~\cite{LAPORTE1992231} solver to generate a shortest path but without returning to the starting point (by setting all edges that return to the starting point with 0 cost). We then route the robot along the path from its initial location to visit the remaining path waypoints.


\subsection{Informative Planning and Online Learning Framework} \label{sec:framework}

For dynamic environment, the prediction accuracy of GP degrades as time elapses because it does not incorporate the temporal variation of the environment. To address this issue, we re-estimate the hyperparameters repetitively at appropriate moments. The re-estimate triggering mechanism depends on two factors:
\begin{itemize}
\item The first factor stems from the computational concern. Since any re-estimate will be immediately followed by a re-planning of the future routing path, and because the computation time for the path planning is much more costly than that of the hyperparameter re-estimate. Thus, an appropriate frequency for the simultaneous re-estimate and re-planning needs to be determined to match the computational constraint. 
\item The second factor relates to the intensity of spatiotemporal variations. 
Since the kernel function that describes two points' spatial relation is an indicator of a GP's prediction capacity, 
thus the repetitive hyperparameter re-estimates of the kernel function should reflect the variation intensity of the environment. 
\end{itemize}

In our implementation, we use a measure, $\rho \in [0, 1]$, to decide the moment for triggering the re-estimate and re-planning processes. 
The measure $\rho$ represents the proportion of samples that are recently added to the current BV-set since last re-estimate.
The hyperparameter re-estimate and path re-planning are carried out if $\rho$ is above certain pre-defined threshold, $\rho_0$. 
Roughly, $\rho_0$ can be defined to be inversely proportional to the computational power and the intensity of environmental variation, and the higher the threshold, the less frequent the re-estimate. 
The whole informative planning and online learning framework is pseudo-coded in Alg.~\ref{algo:onlinePlanning}.
\begin{algorithm} \label{algo:onlinePlanning}
    \caption{Informative Planning and Online Learning}
    Initialize SOGP \\
    \While{true} {
        $\rho = 0$ \quad /* for hyperparameter re-estimate */ \\
        Calculate sampling spots as described in \ref{sec:infoPlanner} \\
        Use Travelling Salesman Problem (TSP) solver to generate a routing path, $P$ \\
        \ForEach{point $\bm{p} \in P$} {
            Do sampling on $\bm{p}$ to get a scalar value $v$\\
            Use $(\bm{p}, v)$ as a training point to update SOGP described in \ref{sec:OnlineGP} and \ref{sec:SparseGP} \\
            \If {$(\bm{p}, v)$ replaces some sample in the BV-set} {
                Increase $\rho$
            }
            \If {$\rho > \rho_0$ } {
                Do hyperparameter re-estimate described in \ref{sec:HyperTraining} \\
                break
            }
        }
    }
\end{algorithm}


\section{Experimental Results}

\begin{figure}[t] 
    \centering
    \subfigure[]
    {\label{fig:truth}\includegraphics[height=1.3in]{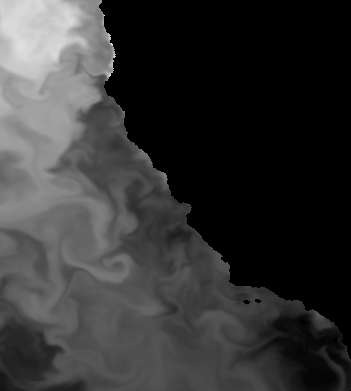}} \quad \quad
    \subfigure[]
    {\label{fig:gp_manual_49obs_annotated}\includegraphics[height=1.3in]{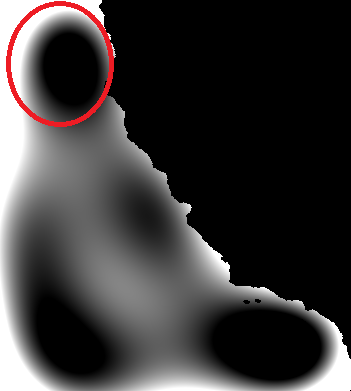}}
    \caption{(a) Salinity data obtained from ROMS. It is treated as a ground truth throughout the paper; (b) The predicted model using GP without data-driven hyperparameter optimization.}
    \label{fig:HyperparametersAccu}
\end{figure}

\begin{figure}[t] 
    \centering
    \subfigure[]
    {\label{fig:DP_4_layer0_0_0obs_manual}\includegraphics[height=1.3in]{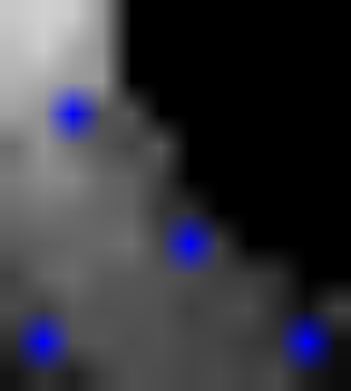}}
    \quad \quad
    \subfigure[]
    {\label{fig:DP_4_layer0_0_0obs_auto}\includegraphics[height=1.3in]{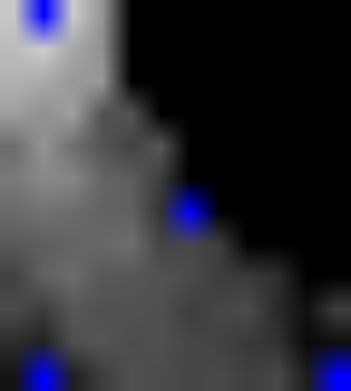}}
    \caption{Informative sampling spots before post-processed as paths. (a) Results under hyperparameters empirically set: 
    $\{\sigma_n^2 = \exp(-2), \sigma_f^2 = \exp(2), l_x = \exp(1), l_y = \exp(1)\}$; 
    (b) Results under hyperparameters learned from data collected: 
    $\{\sigma_n^2 = \exp(-4.6), \sigma_f^2 = \exp(6.8), l_x = \exp(3.4), l_y = \exp(3.2)\}$.
    }
    \label{fig:HyperparametersWaypoints}
\end{figure}

\begin{figure*}[t] 
    \centering
    \subfigure[]
    {\label{fig:1robots_0observations_4DP_round4_lambda1_0}\includegraphics[height=1.2in]{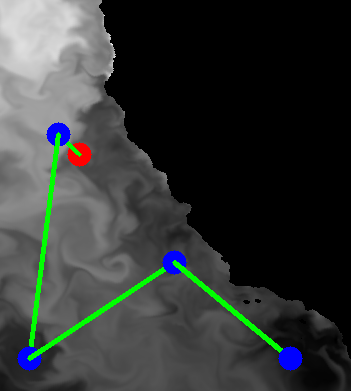}}
    \subfigure[]
    {\label{fig:1robots_0observations_4DP_round4_lambda1_1}\includegraphics[height=1.2in]{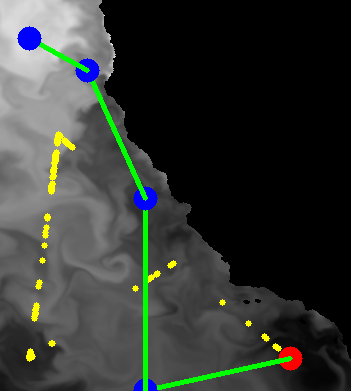}}
    \subfigure[]
    {\label{fig:1robots_0observations_4DP_round4_lambda1_2}\includegraphics[height=1.2in]{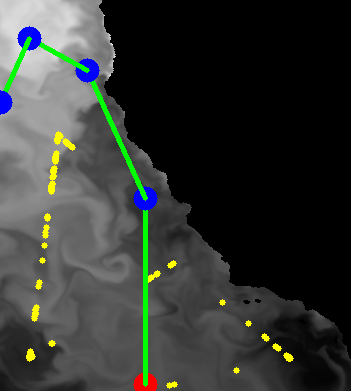}}
    \subfigure[]
    {\label{fig:1robots_0observations_4DP_round4_lambda1_3}\includegraphics[height=1.2in]{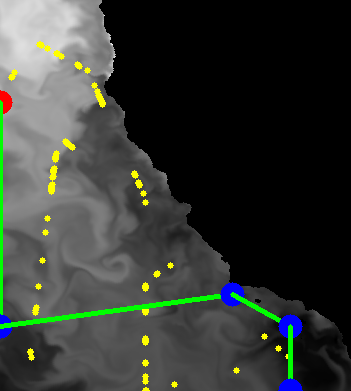}}
    \subfigure[]
    {\label{fig:1robots_0observations_4DP_round4_lambda1_4}\includegraphics[height=1.2in]{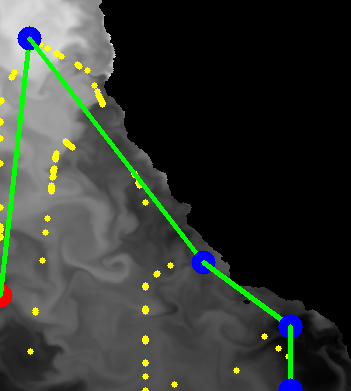}}
    \subfigure[]
    {\label{fig:1robots_0observations_4DP_round4_lambda1_5}\includegraphics[height=1.2in]{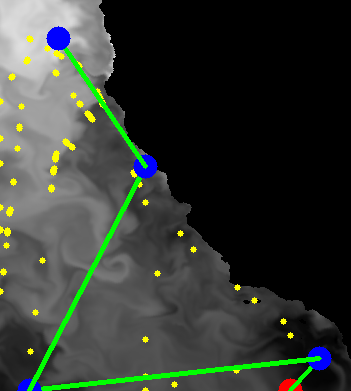}}
    \caption{(a)-(f) Informative paths resulted from subsequent re-plannings. The red and blue points represent the robot's starting locations and the informative sampling spots, respectively. The robot initially launched at (79, 236). The yellow dots denote the points stored in the SOGP BV-set.
}\label{fig:PathsMSE}
\end{figure*}

\begin{figure*}[t]
    \centering
    \subfigure[]
    {\label{fig:sogp_0_stop1}\includegraphics[height=1.2in]{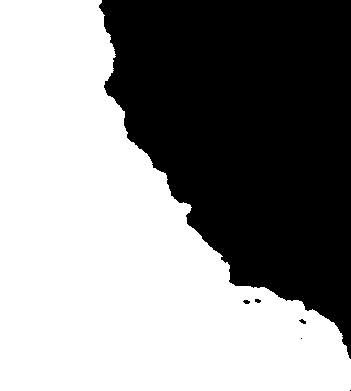}}
    \subfigure[]
    {\label{fig:sogp_510_stop0}\includegraphics[height=1.2in]{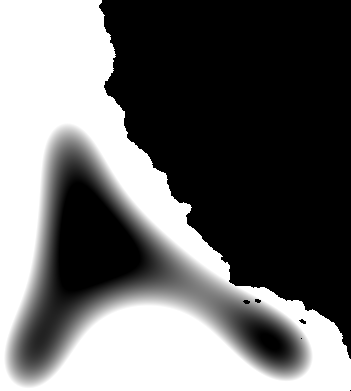}}
    \subfigure[]
    {\label{fig:sogp_663_stop1}\includegraphics[height=1.2in]{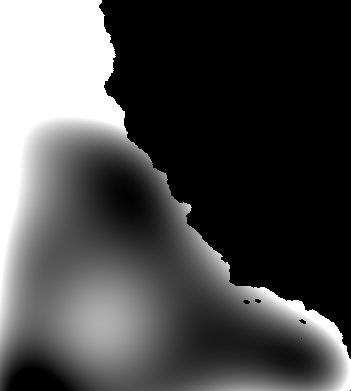}}
    \subfigure[]
    {\label{fig:sogp_1103_stop0}\includegraphics[height=1.2in]{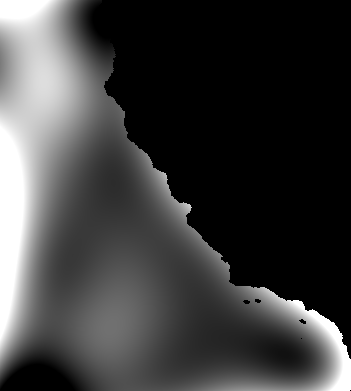}}
    \subfigure[]
    {\label{fig:sogp_1297_stop1}\includegraphics[height=1.2in]{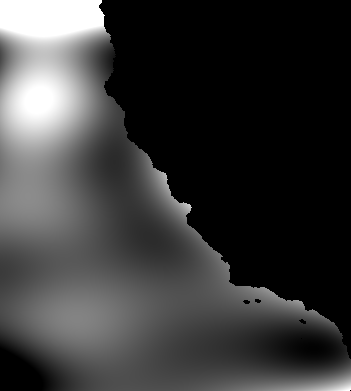}}
    \subfigure[]
    {\label{fig:sogp_1926_stop0}\includegraphics[height=1.2in]{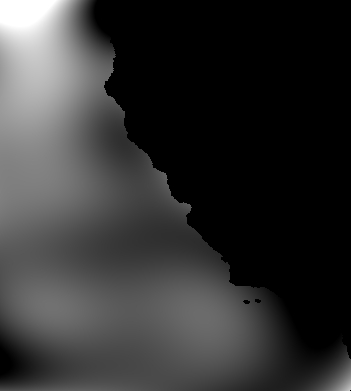}}
    \caption{(a)-(f) The learned environment models. Each corresponds to a step in Fig.~\ref{fig:PathsMSE}.}
     \label{fig:PredictedMap}
\end{figure*}

\begin{figure*}[t] 
    \centering
    {\label{fig:var_2}\includegraphics[height=0.75in]{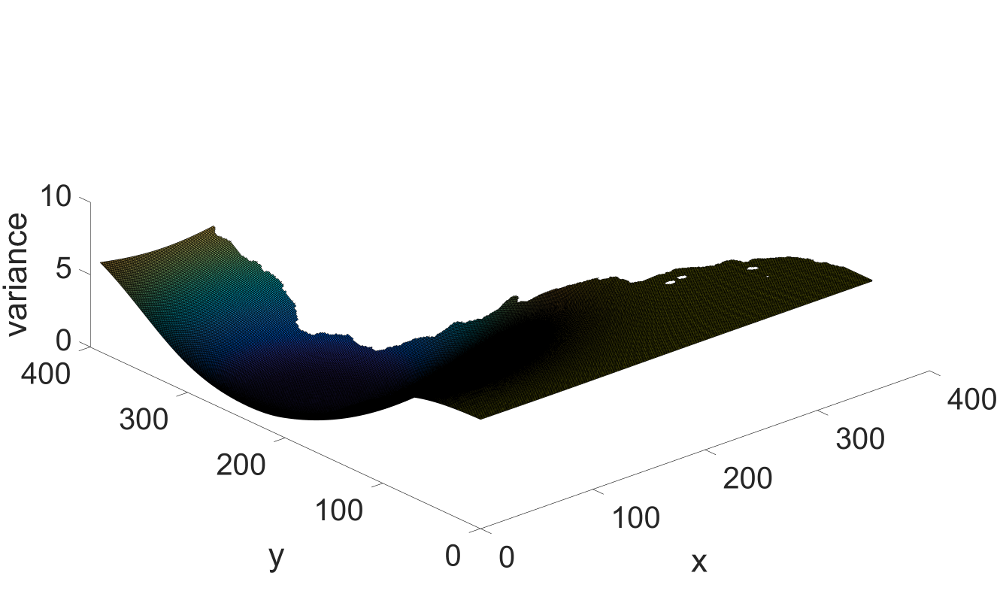}} \ \ 
    \subfigure[]
    {\label{fig:var_3}\includegraphics[height=0.75in]{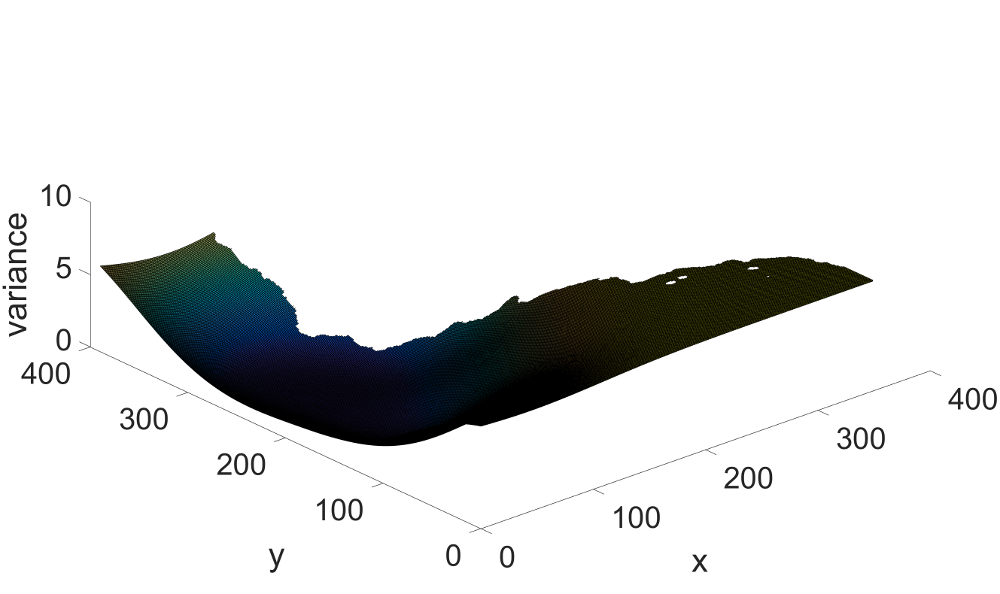}}\ \ 
    \subfigure[]
    {\label{fig:var_4}\includegraphics[height=0.75in]{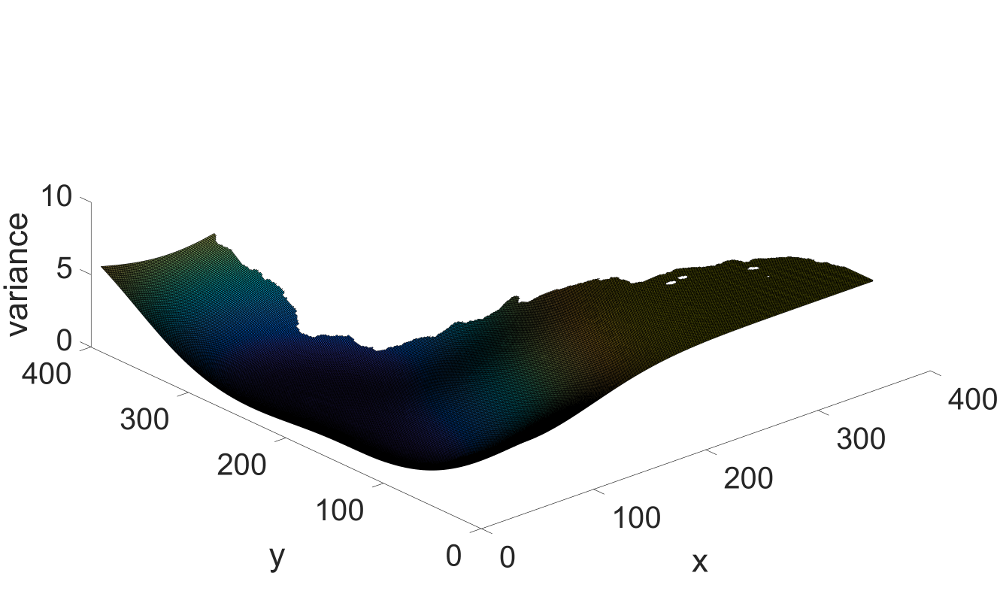}}\ \ 
    \subfigure[]
    {\label{fig:var_5}\includegraphics[height=0.75in]{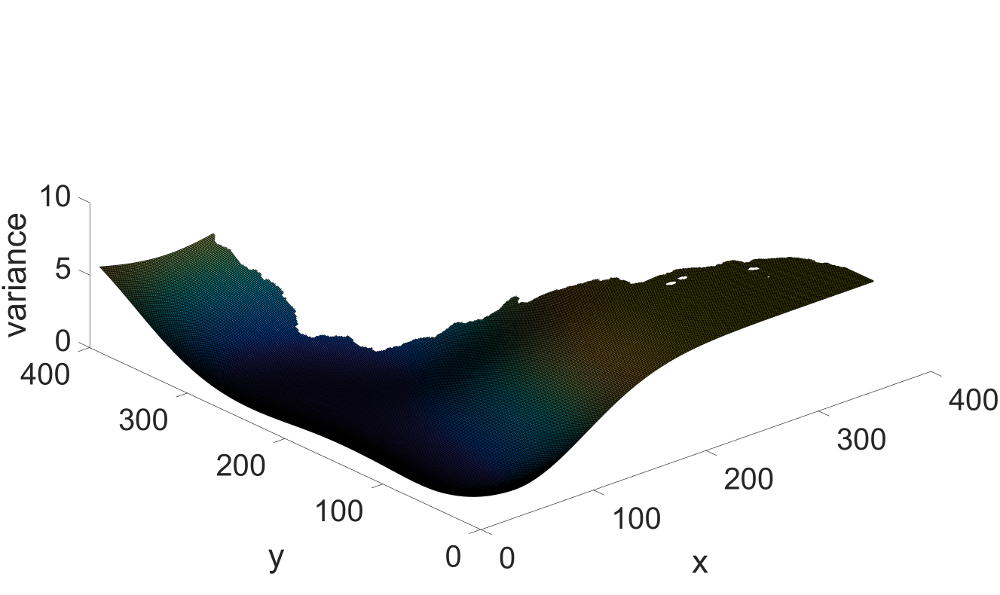}}\ \ 
    \subfigure[]
    {\label{fig:var_6}\includegraphics[height=0.75in]{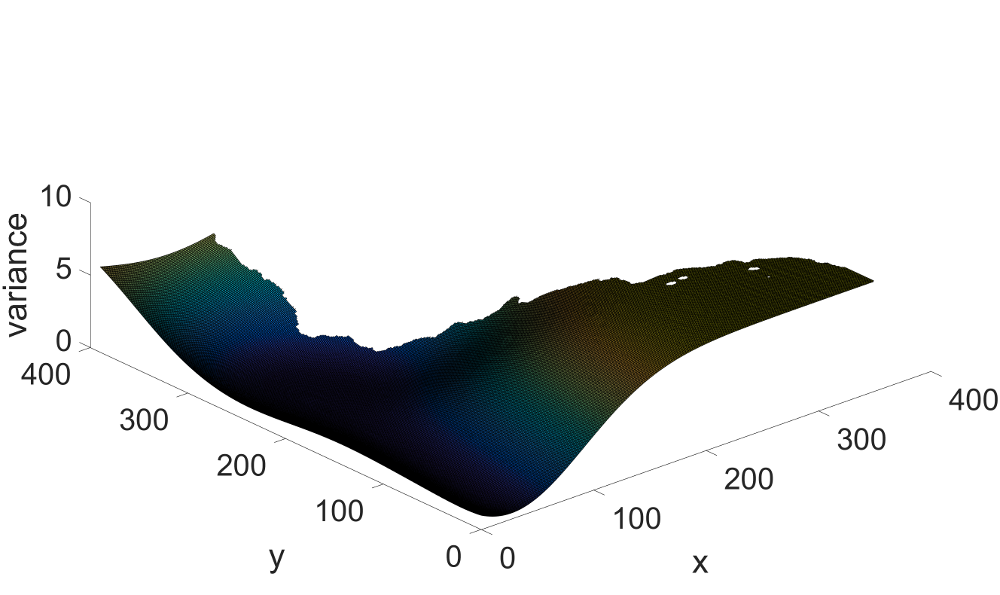}}\ \ 
    \subfigure[]
    {\label{fig:var_7}\includegraphics[height=0.75in]{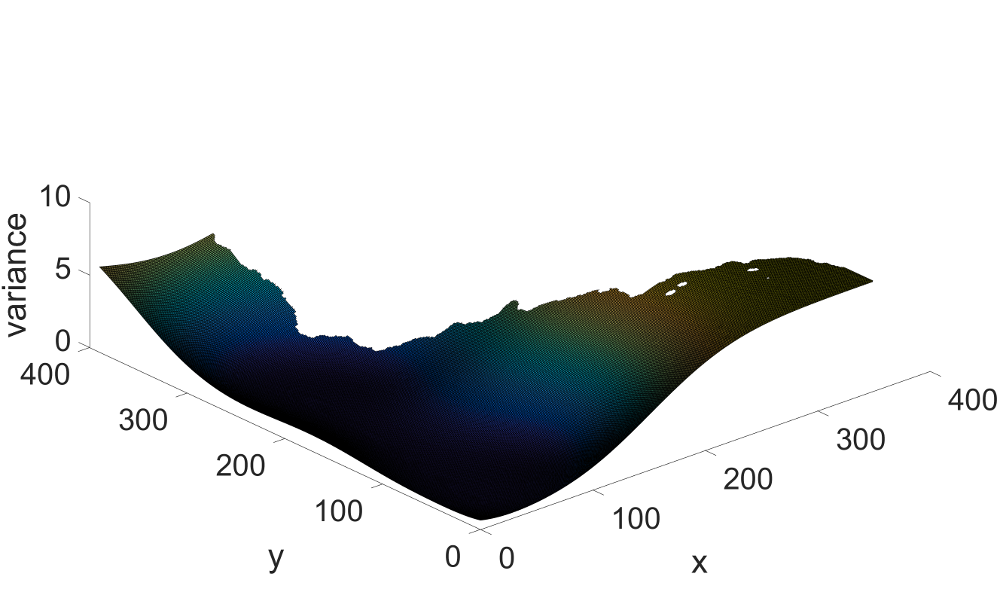}}\ \ 
    \subfigure[]
    {\label{fig:var_8}\includegraphics[height=0.75in]{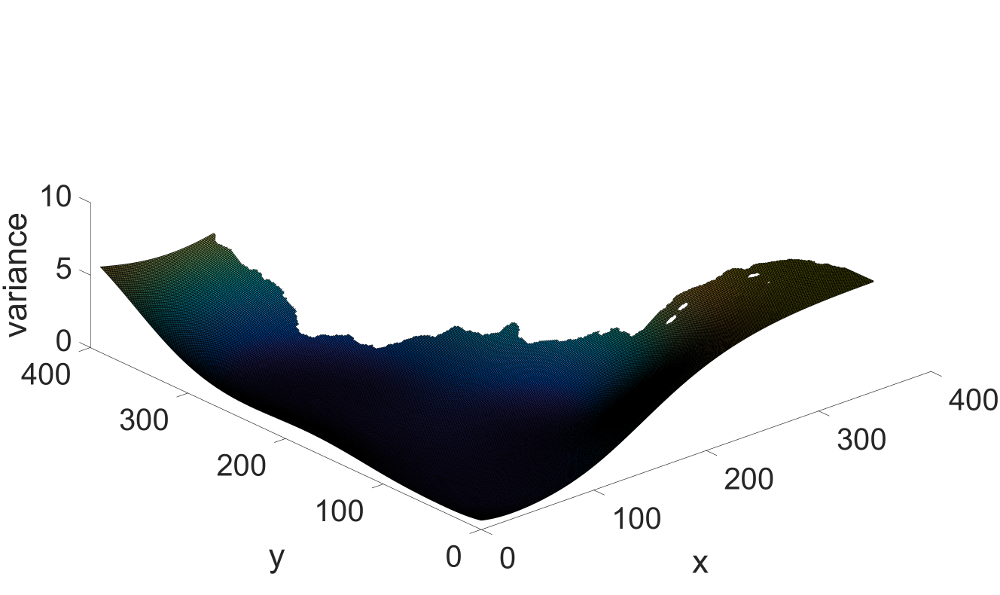}}\ \ 
    \subfigure[]
    {\label{fig:var_9}\includegraphics[height=0.75in]{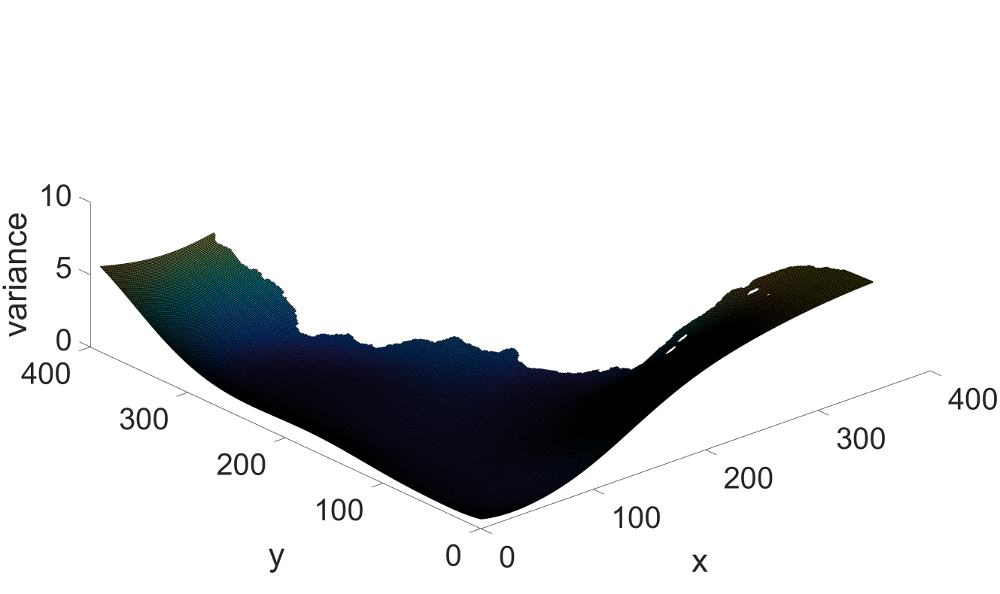}}\ \ 
    \subfigure[]
    {\label{fig:var_10}\includegraphics[height=0.75in]{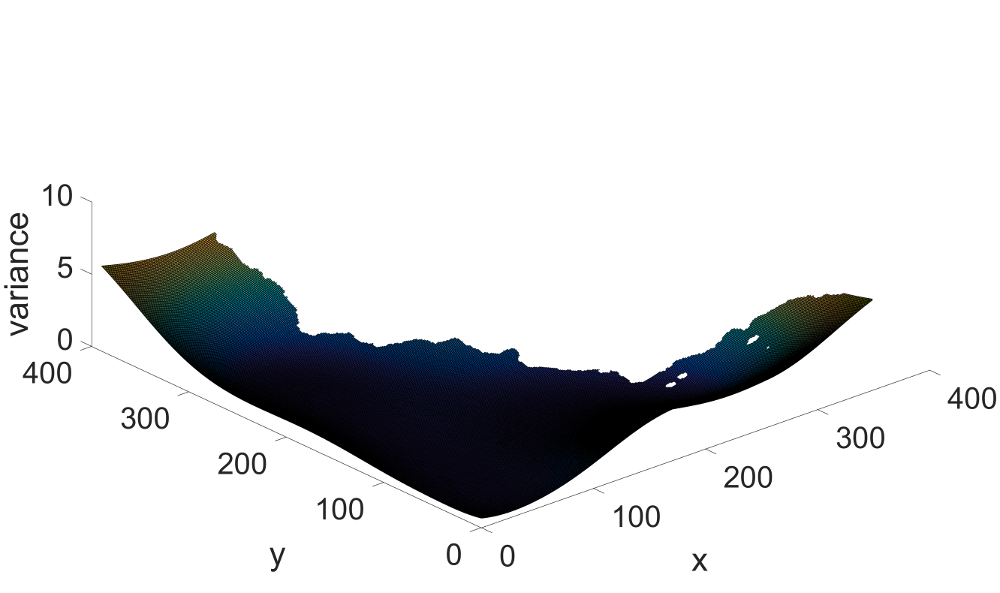}}\ \ 
    \subfigure[]
    {\label{fig:var_last}\includegraphics[height=0.75in]{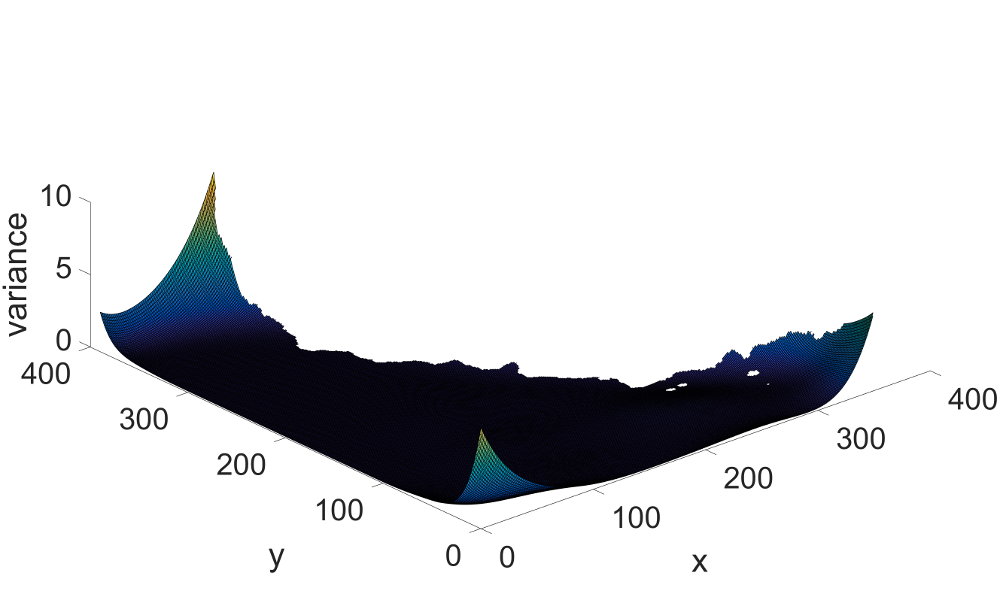}}
    \caption{ Maps of prediction variances. (a)-(i) Variances reduce as the robot follows planned path and collects data samples. (j) The final variance map that corresponds to the moments in Fig.~\ref{fig:1robots_0observations_4DP_round4_lambda1_5} and Fig.~\ref{fig:sogp_1926_stop0}.}
    \label{fig:VarMap}
\end{figure*}

We validated our method in the scenario of ocean monitoring.
The simulation environment was constructed as a two dimensional ocean surface and we tessellated the environment into a grid map. 
Our method applies for any environmental phenomena. In our experiments, we use salinity data recently observed in the Southern California Bight region. The data is obtained from ROMS~\cite{shchepetkin_regional_2005}.
Fig.~\ref{fig:truth} shows the salinity data as a scalar field (the black regions represent lands while the gray areas denote ocean), which is used as the ground truth for comparison.

We implemented a sparse online variant of GP (SOGP) built upon the open-source library {\em libgp}~\cite{libgp}. 
A careful down-sampling of ROMS data to a desired resolution is performed to alleviate the computational cost for generating informative sampling locations. The resolution of the grid map is $351 \times 391$, whereas the resolution for the sampling spots generation (path planning) is $12 \times 12$.

First, we show the predictive accuracy using un-tuned hyperparameters, i.e., hyperparameter values are set empirically/manually instead of data-driven. 
Fig.~\ref{fig:gp_manual_49obs_annotated} shows the prediction result with 50 prior random samples and manually set hyperparameters $\bm{\theta} = \{\sigma_n^2 = \exp(-2), \sigma_f^2 = \exp(2), l_x = \exp(1), l_y = \exp(1)\}$. 
We can observe that the prediction does not match well with the ground truth (see the area circled in red).
Then, we investigate and compare the generated informative sampling points under empirical and data-driven hyperparameters. 
Fig.~\ref{fig:DP_4_layer0_0_0obs_manual} and \ref{fig:DP_4_layer0_0_0obs_auto} show results of
manually-set and data-driven hyperparameters, respectively.
We can see that the relative distances among points (and the covered areas) in Fig.~\ref{fig:DP_4_layer0_0_0obs_auto} are larger than those in Fig.~\ref{fig:DP_4_layer0_0_0obs_manual}. This is mainly affected by $\bm{l}$, which controls the pairwise spatial correlations.

The process of the long-term informative planning and online learning is demonstrated in Fig.~\ref{fig:PathsMSE}.
Each sub-figure depicts an informative path after each hyperparameter re-estimate. 
The red and blue points stand for the robot's current starting position and the informative sampling locations, respectively; the yellow dots represent the points stored in the SOGP BV-set. 
The robot launched from a shore location $(79, 236)$ and performed the sampling operations at each time step along the planned path. 
We emulated the memory limit by setting the maximum size of the BV-set as $m = 100$. The threshold is set as $\rho_0 = 0.6$. 
The distribution patterns of the yellow dots in Fig.~\ref{fig:1robots_0observations_4DP_round4_lambda1_0} to~\ref{fig:1robots_0observations_4DP_round4_lambda1_5} reveal the sparseness of BV-set, indicating that as the robot gradually explores the whole map, the BV-set only stores those points that are the most useful for predicting the model. 
The corresponding prediction maps are shown in Fig.~\ref{fig:PredictedMap}, from which we can see that the constructed models constantly converge to the ground truth and are able to characterize the general patterns of the environment in the final stages.

Finally, we investigate the variances of our predictions. 
We create a variance map on which each value records the variance of a spot on the grid map.  
Fig.~\ref{fig:VarMap} illustrates a series of variance maps along the sampling operations.
We can see that the map gradually ``falls towards the ground",  indicating a decrease of predication variances along the robot's exploration.

\begin{figure}[t]
    \centering
    \subfigure[]
    {\label{fig:79_236}\includegraphics[height=2in]{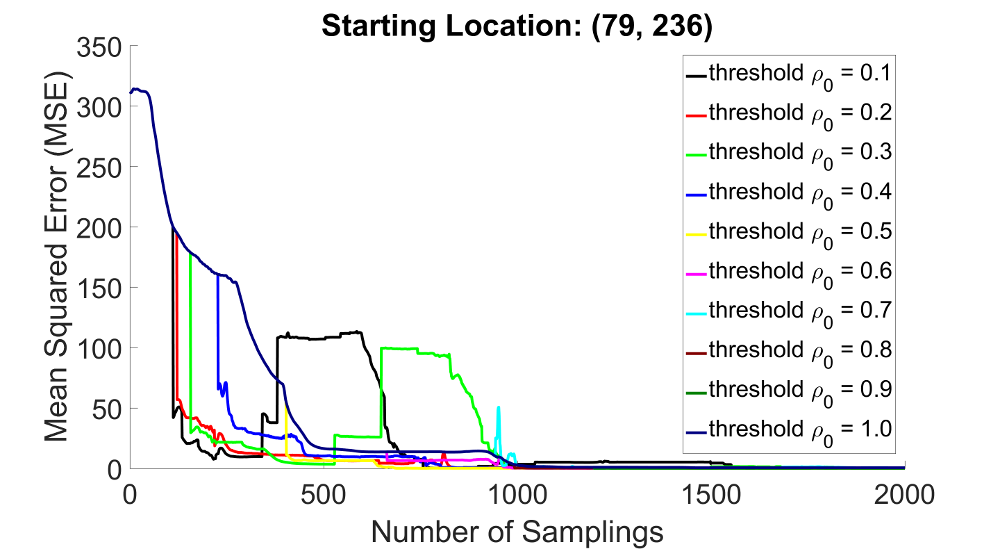}}
    \subfigure[]
    {\label{fig:207_68}\includegraphics[height=2in]{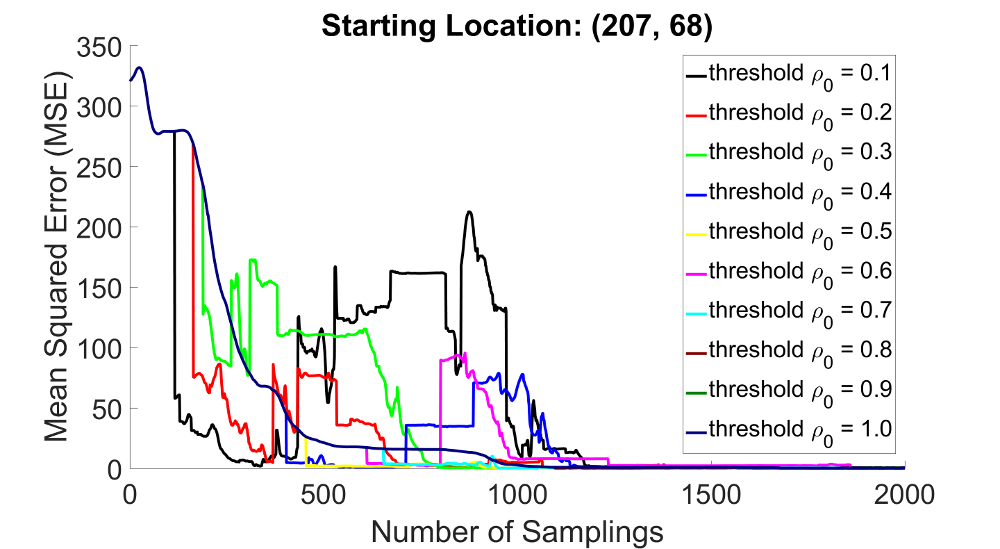}}
    \caption{The MSE plots under different launching locations $\{(79, 236), (207, 68)\}$ and thresholds $\rho_0 = \{0.1, 0.2, \dots, 1.0\}$. The $y$-axis is the MSE value while the $x$-axis is the total number of sampling operations.}
    \label{fig:mse}
\end{figure}

Lastly, Fig.~\ref{fig:mse} shows plots of mean squared errors (MSEs) comparing with the ground truth. We use different thresholds $\rho_0$ and different launch locations to do the statistics.
The $x$-axis corresponds to the total number of sampling operations, which is roughly proportional to the travel time (or distance). 
The $y$-axis is the MSE calculated with the whole map as a testing set. 
The figure reveals that, in general every setting follows a descending trend (reducing error) along the coverage of the planned informative regions. 
By comparing results of different thresholds $\rho_0$, we can observe that there are more error fluctuations for low $\rho_0$ values. A possible reason is that, if the explored regions are not yet well covered, the hyperparameter re-estimate might optimize only among some local regions rather than the entire map, causing a loss of generality and an overfitting problem.

\section{Conclusions}

Environmental monitoring entails persistent presence by robots. This
suggests that both planning and learning are likely to constitute
critical components of any robotic system built for monitoring. In
this paper, we present an informative planning and online learning
method that enables an autonomous marine vehicle to effectively
perform persistent ocean monitoring tasks. Our proposed framework
iterates between a planning component that is designed to collect data
with the richest information content, and a sparse Gaussian Process
learning component where the environmental model and hyperparameters are learned online by
selecting and utilizing only a subset of data that makes the greatest
contribution. We conducted simulations with ocean salinity data;
the results show a good match between the predicted model and the
ground truth, with converging decreases of both prediction errors and map variances.

{ 
\bibliographystyle{abbrv}
\bibliography{reference.bib}
}

\end{document}